\documentclass[sigconf, authorversion=True, nonacm=True]{acmart}
\usepackage{dsfont}
\usepackage{multicol}
\usepackage{multirow}
\usepackage{makecell}
\usepackage{ulem}
\usepackage{enumitem}
\usepackage{balance}

\newcommand{\net}{\mathcal{G}}

\AtBeginDocument{%
  }

\begin{document}

\title{Label Leakage in Federated Inertial-based Human Activity Recognition}

\author{Marius Bock}
\authornote{Both authors contributed equally to this research.}
\email{marius.bock@uni-siegen.de}
\orcid{0000-0001-7401-928X}
\affiliation{%
  \institution{University of Siegen}
  \city{Siegen}
  \country{Germany}
}

\author{Maximilian Hopp}
\authornotemark[1]
\email{maximilian2.hopp@student.uni-siegen.de}
\orcid{0009-0006-3901-4970}
\affiliation{%
  \institution{University of Siegen}
  \city{Siegen}
  \country{Germany}
}

\author{Kristof Van Laerhoven}
\email{kvl@eti.uni-siegen.de}
\orcid{0000-0001-5296-5347}
\affiliation{%
  \institution{University of Siegen}
  \city{Siegen}
  \country{Germany}
}

\author{Michael Moeller}
\email{michael.moeller@uni-siegen.de}
\orcid{0000-0002-0492-6527}
\affiliation{%
  \institution{University of Siegen}
  \city{Siegen}
  \country{Germany}
}

\renewcommand{\shortauthors}{Marius Bock, Maximilian Hopp, Kristof Van Laerhoven, and Michael Moeller}

\begin{abstract}
 While prior work has shown that Federated Learning updates can leak sensitive information, label reconstruction attacks, which aim to recover input labels from shared gradients, have not yet been examined in the context of Human Activity Recognition (HAR). Given the sensitive nature of activity labels, this study evaluates the effectiveness of state-of-the-art gradient-based label leakage attacks on HAR benchmark datasets. Our findings show that the number of activity classes, sampling strategy, and class imbalance are critical factors influencing the extent of label leakage, with reconstruction accuracies reaching well-above 90\% on two benchmark datasets, even for trained models. Moreover, we find that Local Differential Privacy techniques such as gradient noise and clipping offer only limited protection, as certain attacks still reliably infer both majority and minority class labels. We conclude by offering practical recommendations for the privacy-aware deployment of federated HAR systems and identify open challenges for future research. 
 Code to reproduce our experiments is publicly available via \url{github.com/mariusbock/leakage_har}.
\end{abstract} 




\maketitle

\section{Introduction}
\label{sec:intro}

Federated training of deep learning classifiers has received increasing attention in recent years \cite{bonawitzFederatedLearningScale2019, yangAppliedFederatedLearning2018, dayanFederatedLearningPredicting2021}. In scenarios involving sensitive data, Federated Learning (FL) enables multiple users to collaboratively train a global model while keeping their local data private \cite{mcmahanCommunicationEfficientLearningDeep2017}. By exchanging only model updates, such as gradients or locally trained weights, FL was initially considered to offer strong privacy guarantees. However, subsequent research has demonstrated that these updates may still leak sensitive information \cite{phongPrivacyPreservingDeepLearning2017, phongPrivacyPreservingDeepLearning2018}, including class labels \cite{geipingInvertingGradientsHow2020, wainakhLabelLeakageGradients2021, wainakhUserLevelLabelLeakage2022,zhaoIDLGImprovedDeep2020, gatHarmfulBiasGeneral2024, maInstancewiseBatchLabel2023} and even reconstructed representations of the original input data \cite{zhuDeepLeakageGradients2019, geipingInvertingGradientsHow2020, yinSeeGradientsImage2021, zhuRGAPRecursiveGradient2021}. With the automatic recognition of activities through wearable devices such as smartwatches having emerged as a valuable tool for numerous applications, large-scale centralized approaches to Human Activity Recognition (HAR) face considerable challenges due to privacy concerns and legal constraints, and communication inefficiencies \cite{liSurveyFederatedLearning2025}. As a result, by allowing only model updates to be shared, Federated Learning for Human Activity Recognition (FL-HAR) has become a promising new approach to HAR to safeguard both sensor and label information on user devices \cite{sozinovHumanActivityRecognition2018, yuFedHARSemiSupervisedOnline2023, ekFederatedLearningAggregation2021, liHierarchicalClusteringbasedPersonalized2023}. 

Although prior work has shown that information leakage can occur through FL updates, limited work has addressed these threats specifically within HAR environments \cite{presottoPreliminaryResultsSensitive2022, royPrivateFairSecure2023, xiaoFederatedLearningSystem2021, liuFederatedPersonalizedRandom2021, kerkoucheClientspecificPropertyInference2023, elhattabPASTELPrivacyPreservingFederated2023, chenPrivateDataLeakage2024}. In particular, gradient inversion attacks, which attempt to reconstruct input data and associated labels from shared gradients, have not yet been examined in the context of HAR. Given that activity labels may reveal highly sensitive user behaviors and personal activity patterns, this study investigates the applicability of state-of-the-art, gradient-based label leakage attacks, originally developed for computer vision tasks, when applied to HAR benchmark datasets.

Our contributions are three-fold:
\begin{enumerate}
    \item We conduct a comprehensive evaluation of five gradient inversion techniques for label reconstruction across two widely used HAR datasets and model architectures. We also assess the effectiveness of local defense mechanisms in mitigating such attacks.
    \item We demonstrate that both the number of activity classes in a dataset, degree of imbalance and the sampling strategy used during training play critical roles in determining the extent of label leakage, even for fully trained HAR models.
    \item We show that the unbalanced nature of HAR datasets introduces unique privacy risks in federated settings, and we assess how this necessitates stronger applications of Local Differential Privacy (LDP) methods to effectively hide user label information.
\end{enumerate}

\section{Related Work}
\label{sec:related}

Body-worn sensor data, particularly in the context of HAR, involves the collection of sensitive user information. As such, it is typically subject to strict data privacy regulations and is not easily shareable in centralized learning environments \cite{liSurveyFederatedLearning2025}. In response, researchers have explored federated learning as a privacy-conscious approach to training HAR models \cite{sozinovHumanActivityRecognition2018, yuFedHARSemiSupervisedOnline2023, ekFederatedLearningAggregation2021, liHierarchicalClusteringbasedPersonalized2023}. Given that related fields like computer vision have revealed potential vulnerabilities in FL environments, researchers have also begun investigating similar risks in HAR scenarios \cite{presottoPreliminaryResultsSensitive2022, royPrivateFairSecure2023, xiaoFederatedLearningSystem2021, liuFederatedPersonalizedRandom2021}. One of the earliest works in this domain by Presotto et al. \cite{presottoPreliminaryResultsSensitive2022} demonstrated the feasibility of Membership Inference Attacks (MIA) that determine whether a user participated in the training process of a federated model. Subsequent studies by Kerkouche et al. \cite{kerkoucheClientspecificPropertyInference2023}, Elhattab et al. \cite{elhattabPASTELPrivacyPreservingFederated2023}, and Chen et al. \cite{chenPrivateDataLeakage2024} expanded upon Presotto et al.’s findings, further investigating MIA vulnerabilities in FL-HAR systems. Roy et al. \cite{royPrivateFairSecure2023} additionally explored the integration of local differential privacy measures while maintaining fairness in client update selection.

Early research into data reconstruction from gradients was pioneered by Phong et al. \cite{phongPrivacyPreservingDeepLearning2017, phongPrivacyPreservingDeepLearning2018}. Later, Zhu et al. \cite{zhuDeepLeakageGradients2019} and Geiping et al. \cite{geipingInvertingGradientsHow2020} introduced attacks capable of reconstructing original input data from gradients in FL systems. Beyond input reconstruction, more recent work has focused on label leakage, i.e., inferring label information such as class presence or batch label distributions from gradient updates \cite{gatHarmfulBiasGeneral2024, wainakhLabelLeakageGradients2021, maInstancewiseBatchLabel2023, yinSeeGradientsImage2021, zhaoIDLGImprovedDeep2020}. To date, label leakage attacks have not been investigated in the context of HAR. Given that sensor-based HAR data and its use cases differ fundamentally from those in computer vision, our study aims to demonstrate how these differences, such as the prevalence of a majority NULL class, temporal consistency across records, and small number of classes, give rise to distinct privacy risks in FL-HAR systems.

\begin{figure}
    \centering
    \includegraphics[width=0.85\linewidth]{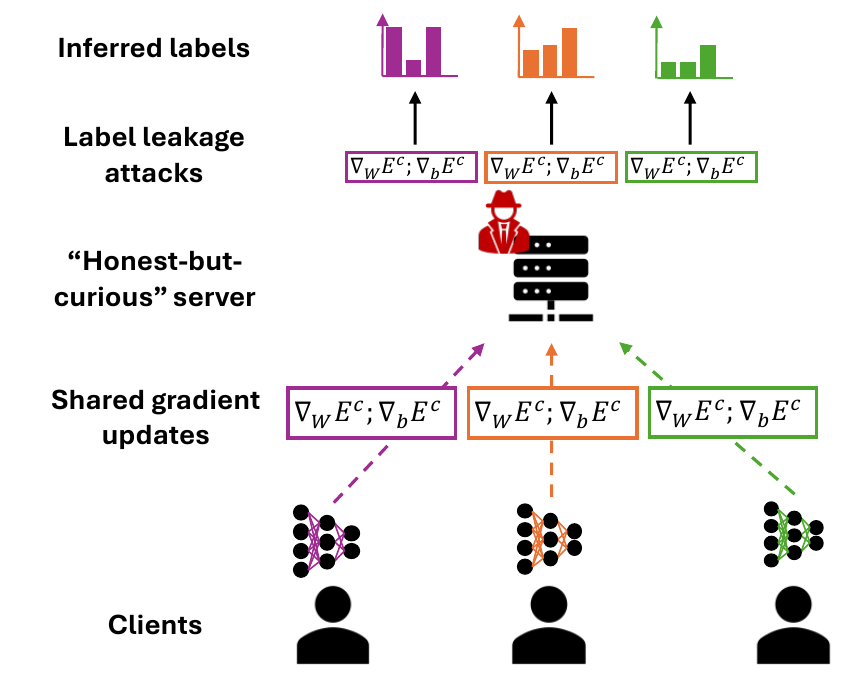}
    \caption{Overview of the applied threat model. An \textit{honest-but-curious server} exploits gradient updates received from individual clients to infer the presence and distribution of class labels within local batches. Label leakage attacks are conducted by analyzing the gradients of the weights and biases associated with the final layer $L$ of the trained model.}
    \label{fig:threatmodel}
    \Description{Overview of the applied threat model. An \textit{honest-but-curious server} exploits gradient updates received from individual clients to infer the presence and distribution of class labels within local batches. Label leakage attacks are conducted by analyzing the gradients of the weights and biases associated with the final layer $L$ of the trained model.}
\end{figure}
\section{Methodology}
\label{sec:methodolgy}

\subsection{Problem Setting \& Threat Model}
\label{subsec:problem}

We study a gradient-based FL setting in which multiple clients $C$ collaboratively train a global neural network $G$. A central server $S$ coordinates the training by aggregating client-submitted gradients and updating the global model using gradient descent. More specifically, each client $c$ receives a snapshot $\theta^S:=(W^{S},b^S)$ of the current weights and biases of a model $\net$ (summarized into model parameters $\theta^S$) by the server, constructs a cost function 
\begin{align}
    E^c(\theta) =\sum_{(x_i,y_i) \in T^c} \mathcal{L}(\net(x_i;\theta), y_i),
\end{align}
based on local client training data $(x_i,y_i) \in T^c$, a suitable loss function $\mathcal{L}$, and performs one or multiple steps of gradient descent on $E_c$. The new parameters are subsequently sent to the server. In the simplest case of a single gradient descent step, the data sent to the server are the gradients $(\nabla_W E^c,\nabla_b E^c)$ of the costs w.r.t. the weights and biases of the network. In this work, we study what this information reveals about the empirical distribution $\hat{p}(y)$ of the activity labels $y_i$ in the client's training data $T^c$. Specifically, we focus on the reconstruction of label information from gradients of the last layer $L$ through gradient inversion attacks, as detailed in Section~\ref{subsec:attacks}. Throughout our work, we assume that clients compute and share correct gradients without using class weights based on valid data and labels and also assume that the server knows the number of training examples $|T^c| =:N$, which were used to compute the individual gradients. The threat model considered is that of an \textit{honest-but-curious server}: the server $S$ performs protocol-compliant aggregation but is passively interested in extracting sensitive information from the received updates (see Figure~\ref{fig:threatmodel}). While alternative threat models, such as an actively malicious server that modifies the model to enhance information leakage, or clients that train with more than $N$ data points, are plausible, they fall outside the scope of this work. 

\subsection{Label Reconstruction Attacks}
\label{subsec:attacks}
We consider both weight-based and bias-based attacks, which respectively exploit the gradients of the weights ($\nabla_W E^c$) and biases ($\nabla_b E^c$) of the model’s final layer, i.e., the classification layer. All attack methods operate under a white-box assumption, wherein the server has full knowledge of the model architecture and the batch size used to compute each gradient update. We consider these assumptions realistic, given that the server orchestrates and oversees the federated learning process.

\paragraph{Weight-based Attacks}

For weight-based label reconstruction, we employ the LLG and LLG* attacks introduced by Wainakh et al.~\cite{wainakhUserLevelLabelLeakage2022}. These methods exploit two key properties of the weight gradients in the final (classification) layer. First, for a given class label $i$, the sum of the corresponding weight gradient $(\nabla_{W_i} E^c)$ tends to be negative if label $i$ is present in the local training batch. Second, in an untrained model, the total contribution of class $i$ to the weight gradient is approximately proportional to the number of samples in the batch with that label. Based on this, the authors define a constant $m$ representing the gradient impact of a single instance of label $i$. The LLG* variant assumes additional attacker knowledge of the model architecture and parameters, allowing for more accurate estimation of $m$ using synthetic batches of dummy data.

\paragraph{Bias-based Attacks}

The LLBG attack, proposed by Gat et al.\cite{gatHarmfulBiasGeneral2024}, consists of a two-phase procedure. In the first phase, all labels that are guaranteed to be present in the batch, i.e. having a negative bias gradient component $g_i$, are added to the reconstructed label set. The associated gradient values are then increased by an estimated $m$, similar to the LLG approach \cite{wainakhUserLevelLabelLeakage2022}. They estimate that for untrained models the impact of a class is defined by $\beta_i \approx -\frac{\lambda_i}{N}$, where $\lambda_i$ is the number of samples of class $i$ in the batch, $N$ is the batch size and $\beta_i$ is a single component of the bias gradient, thus LLBG uses $-1/N$ as $m$ for untrained models.
In the second phase, the label $i$ corresponding to the current minimum bias gradient component $g_i$ is iteratively appended to the reconstructed list, with its value increased by $m$. If the gradient remains minimal, label $i$ is again added to the reconstructed list; otherwise, the process switches to the new minimum component in $g$. This continues until the reconstructed label set reaches batch size $N$.

The EBI attack is a baseline that is used in~\cite{gatHarmfulBiasGeneral2024} tries to estimate the impact $m$ of a label empirically by computing 
\begin{align}
    m = \frac{1}{N} \sum_{i;\,\beta_i < 0} \beta_i,
\end{align}
where $N$ is the batch size, $\beta$ is the complete bias gradient $(\nabla_{b} E^c)$ and $i$ is used to indicate a specific component of $\beta$.
In that regard, it is similar to the LLG attack \cite{wainakhUserLevelLabelLeakage2022} but uses the algorithm of LLBG \cite{gatHarmfulBiasGeneral2024}. While a variant of LLBG tailored for trained models is proposed in~\cite{gatHarmfulBiasGeneral2024}, it requires knowledge of model prediction accuracy and is thus excluded from this work. The iLRG attack by Ma et al.~\cite{maInstancewiseBatchLabel2023} formulates label reconstruction as solving a system of linear equations over batch-averaged gradients. 

\subsection{Defense Measures}
Though research has continuously worked on privacy-enhancing methods such as Differential Privacy (DP) \cite{mcmahanLearningDifferentiallyPrivate2018} and encryption-based techniques \cite{bonawitzFederatedLearningScale2019}, many of these methods require coordination of the server, i.e. requiring trust of the users in the server. The Federated Learning environment sketched in Figure \ref{fig:threatmodel} assumes that users do not trust the server they are sending their gradient-based updates to. We thus only focus on Local Differential Privacy methods, namely gradient clipping and noise addition. These methods, being executed locally by each user individually, do not require users to trust the server. In a realistic setting, a privacy budget $\epsilon$ determines the degree of LDP to be applied during the federated training process of the neural network. This work, though, applies the approach of Gat et al. \cite{gatHarmfulBiasGeneral2024} where no privacy budget needs to be retained because the label reconstruction is evaluated on single batches sampled from the datasets without running a complete training scenario. The clipping of the gradients is done by normalizing the gradient vectors to an L2-norm of $\rho$ if the norm of the gradient exceeds a defined threshold. Noise is added to the gradients by sampling  from a Gaussian distribution that has a mean of $\mu = 0$ and a predefined standard deviation of $\sigma$.

\subsection{Experimental Setup}
\label{subsec:experiments}
We investigate label leakage in HAR using two widely adopted lightweight deep learning architectures DeepConvLSTM \cite{ordonezDeepConvolutionalLSTM2016} and TinyHAR \cite{zhouTinyHARLightweightDeep2022}. The DeepConvLSTM model comprises a sequence of four convolutional layers, two LSTM layers, and a final classification layer. TinyHAR extends this design by incorporating optimized temporal feature extraction mechanisms, such as self-attention, while also significantly reducing the parameter count, thereby enhancing its suitability for deployment on resource-constrained edge devices. For evaluation, we utilize two real-world HAR datasets, namely the WEAR \cite{bockWEAROutdoorSports2024} and Wetlab dataset \cite{schollWearablesWetLab2015}. Both datasets include a NULL-class representing periods not associated with any target activity. The WEAR dataset features participants performing 18 distinct sports activities outdoors while wearing four inertial measurement units (IMUs) on their limbs and a head-mounted camera. In our experiments, we use a pre-release version of the dataset consisting of 18 participants and rely solely on IMU-data. The Wetlab dataset includes 22 participants executing DNA extraction procedures from onions and tomatoes in a laboratory setting, while wearing a wrist-mounted IMU on the dominant hand. This dataset comprises 9 activity classes, such as cutting, inverting, and peeling. 
For a specific dataset and a specific client with local training data $T^c$, we define the histogram
\begin{align}p^{\text{gt}} := \sum_{y_i \in T^c} e_{y_i}\end{align}
for $e_j$ being the $j-th$ unit vector. We compute an estimate $p$ of $p^{\text{gt}} $ using one of the gradient inversion attacks and evaluate
\begin{itemize}
    \item the \textit{Label Existence Accuracy (LeAcc)}, 
    \begin{align}
    \frac{1}{|n|}\sum_{j=1}^n \left|(p^{\text{gt}}_j>0) - (p_j>0)\right|,
    \end{align}
    where the $>0$  denotes a thresholding,
    \item the \textit{Label Number Accuracy (LnAcc)}, 
    \begin{align}
    \frac{1}{|T^c|}\sum_{j=1}^n \min(p^{\text{gt}}_j, p_j)
    \end{align}
\end{itemize}
and average these results over all $T^c$ of a client to compute a single LeAcc and LeAcc metric. As LeAcc and LnAcc are compute per batch, datasets which show an unbalanced class count can cause these metrics to be biased towards majority classes. We thus propose a third metric, \textit{Class-wise Average Accuracy (ClassAcc)}, which is calculated as the average of the per-class LnAcc computed across the complete dataset of a client, i.e. all $T^c$ that were reconstructed of a specific client.

\section{Results}
\label{sec:results}
\begin{figure*}
    \centering
    \includegraphics[width=0.9\linewidth]{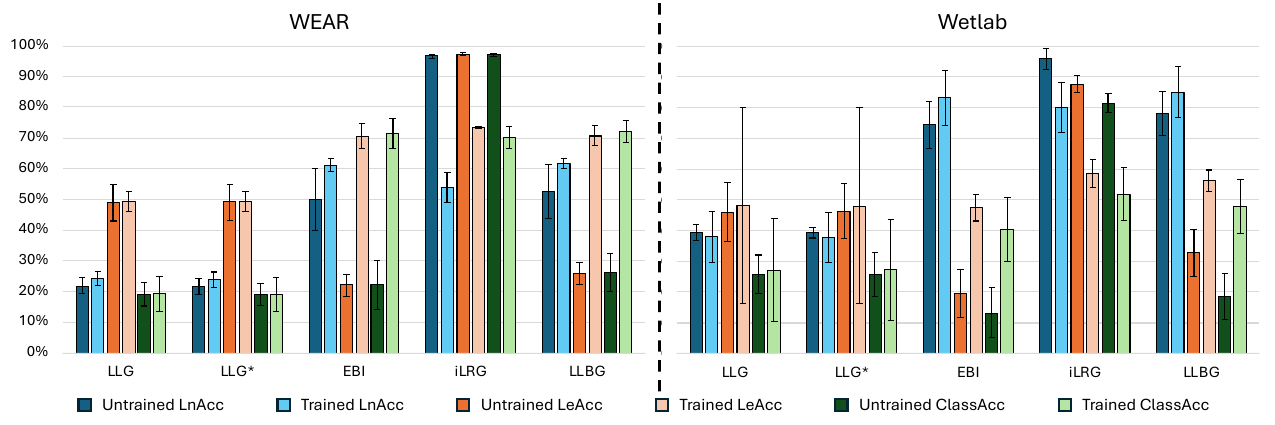}
    \caption{Comparison of LnAcc, LeAcc and ClassAcc of the investigated attacks using a shuffling sampling technique and a batch size of 100 samples. Each value is the average across the different dataset (WEAR and Wetlab) and architecture combinations. Error bars refer to Standard Deviation (SD) across clients.}
    \label{fig:overview}
    \Description{Comparison of LnAcc, LeAcc and ClassAcc of the investigated attacks using a shuffling sampling technique and a batch size of 100 samples. Each value is the average across the different dataset (WEAR and Wetlab) and architecture combinations. Error bars refer to Standard Deviation (SD) across clients.}
\end{figure*}

The following presents the experimental results of each label leakage attack across all combinations of datasets and model architectures. We examine the influence of different sampling strategies, model states (trained vs. untrained), single- versus multi-step updates and the effectiveness of privacy-preserving mechanisms. All experiments were conducted using an unweighted cross-entropy loss, applying the leakage attacks on the batch-wise gradients returned by the last layer of the respective model. 

Both the WEAR \cite{bockWEAROutdoorSports2024} and Wetlab datasets \cite{schollWearablesWetLab2015} are sampled at 50Hz, and input sequences are generated using a sliding window of 1 second (50 samples) with a 50\% overlap. Label leakage for each client, i.e. participant, is evaluated using a Leave-One-Subject-Out cross-validation, which means that data of the client for which label leakage is to be investigated was previously unseen to the global model. Note that a \textit{trained} model refers to a model which has been locally trained by the server for 100 epochs using all other participants data using the training setup as reported in \cite{bockTemporalActionLocalization2024}. Models are assessed by how well they reconstruct the local data $T^c$ of clients, iterating through the clients data in batch-wise manner with the batch size being the size of $T^c$.

\paragraph{Single-step gradient updates}
\label{p:single-step}
Figure \ref{fig:overview} presents the results of the evaluated label leakage attacks on the WEAR and Wetlab datasets, using shuffled sampling with a batch size of 100 sliding windows. Overall, bias-based techniques, namely EBI, iLRG, and LLBG, consistently outperform the weight-based techniques LLG and LLG* by a significant margin. Among all methods, iLRG achieves the highest performance on untrained models, with an average ClassAcc of approximately 97\% on the WEAR dataset and 81\% on the Wetlab dataset. In contrast, for trained models, EBI and LLBG also become effective attacks, achieving around 70\% on WEAR and between 40-47\% on Wetlab. In addition to the shuffled sampling approach, we evaluate two alternative sampling techniques: sequential and balanced. Sequential sampling preserves the temporal order of sliding windows, while balanced sampling constructs batches containing an equal number of samples per class by randomly drawing from class-specific pools. Note that to address class imbalance in the dataset, balanced batches are filled using repeated samples as needed. 

Table \ref{tab:sampling_techniques} reports the average performance for each sampling technique, aggregated across bias- and weight-based attacks and model architectures. Results show that sequentially sampled batches exhibit significantly more label leakage than balanced or shuffled batches. When comparing the performance of attacks on untrained versus trained models, LLBG and EBI show improved performance on trained models when using shuffled and balanced sampling. In contrast, iLRG experiences a noticeable drop in effectiveness when applied to trained models. Among all methods, EBI remains the most stable across both trained and untrained scenarios. When comparing the two architectures, TinyHAR and DeepConvLSTM, performance varies across experiments. However, a trained TinyHAR model shows greater vulnerability to the iLRG attack, with higher leakage levels across all evaluations. Meanwhile, EBI and LLBG generally perform better on the DeepConvLSTM than on TinyHAR when models are trained. Looking at dataset differences, the WEAR dataset consistently exhibits greater label leakage than Wetlab. In particular under sequential sampling, EBI and LLBG maintain ClassAcc levels above 90\% on WEAR, even with trained models. This heightened vulnerability is likely due to the recording setup of the WEAR dataset, which contains fewer activity transitions. As a result, sequential batches more frequently become label-exclusive, i.e., containing only a single class. Our results indicate that such batches pose a serious privacy risk, as labels can be accurately reconstructed in various settings. In summary, since shuffled batches consistently yield the lowest leakage, we hypothesize that effective mitigation of label leakage in HAR involves constructing batches that contain a diverse set of activity classes, yet are not fully class-balanced.

\begin{table}
    \centering
    \footnotesize
    \begin{tabular}{ll|cccc} 
        \multicolumn{2}{l}{Sampling Technique} & WEAR (U)  & WEAR (T) & Wetlab (U)  & Wetlab (T) \\ \hline 
        \multirow{6}{*}{Shuffle}    & \multirow{2}{*}{Weight}  & 19.09\%     & 19.13\%      & 25.89\%     & 27.12\% \\
                                    &                          & ($\pm3.67$) & ($\pm5.66$)  & ($\pm6.69$) & ($\pm16.63$) \\
                                    & \multirow{2}{*}{Bias}    & 48.50\%     & 71.21\%      & 37.76\%     & 46.67\% \\
                                    &                          & ($\pm4.89$) & ($\pm4.02$)  & ($\pm6.25$) & ($\pm9.23$) \\
                                    & \multirow{2}{*}{Avg.}    & 36.74\%     & 50.38\%      & 33.01\%     & 38.85\% \\
                                    &                          & ($\pm4.40$) & ($\pm4.68$)  & ($\pm6.42$) & ($\pm12.19$) \\ \hline
        \multirow{6}{*}{Sequential} & \multirow{2}{*}{Weight}  & 27.26\%     & 40.32\%      & 25.05\%     & 30.75\% \\
                                    &                          & ($\pm4.47$) & ($\pm32.70$) & ($\pm5.81$) & ($\pm18.89$) \\
                                    & \multirow{2}{*}{Bias}    & 98.64\%     & 80.71\%      & 81.67\%     & 70.76\% \\
                                    &                          & ($\pm0.66$) & ($\pm3.67$)  & ($\pm4.74$) & ($\pm6.44$) \\
                                    & \multirow{2}{*}{Avg.}    & 70.09\%     & 64.56\%      & 59.02\%     & 54.75\% \\
                                    &                          & ($\pm2.18$) & ($\pm15.28$) & ($\pm5.17$) & ($\pm11.42$) \\ \hline
        \multirow{6}{*}{Balanced}   & \multirow{2}{*}{Weight}  & 14.34\%     & 14.98\%      & 15.65\%     & 15.91\% \\
                                    &                          & ($\pm0.37$) & ($\pm4.81$)  & ($\pm1.07$) & ($\pm2.79$) \\
                                    & \multirow{2}{*}{Bias}    & 70.07\%     & 71.00\%      & 68.33\%     & 65.94\% \\
                                    &                          & ($\pm0.60$) & ($\pm2.45$)  & ($\pm2.47$) & ($\pm6.26$) \\
                                    & \multirow{2}{*}{Avg.}    & 47.77\%     & 48.59\%      & 47.26\%     & 45.93\% \\
                                    &                          & ($\pm0.51$) & ($\pm3.39$)  & ($\pm1.91$) & ($\pm8.56$) \\
    \end{tabular}
    \caption{Average ClassAcc of bias- and weight-based attacks applied to WEAR and Wetlab using shuffled, sequential and balanced sampling. We differentiate between trained (T) and untrained (U) models and report SD across clients.}
    \label{tab:sampling_techniques}
\end{table}

\begin{figure}
    \centering
    \includegraphics[width=1.0\linewidth]{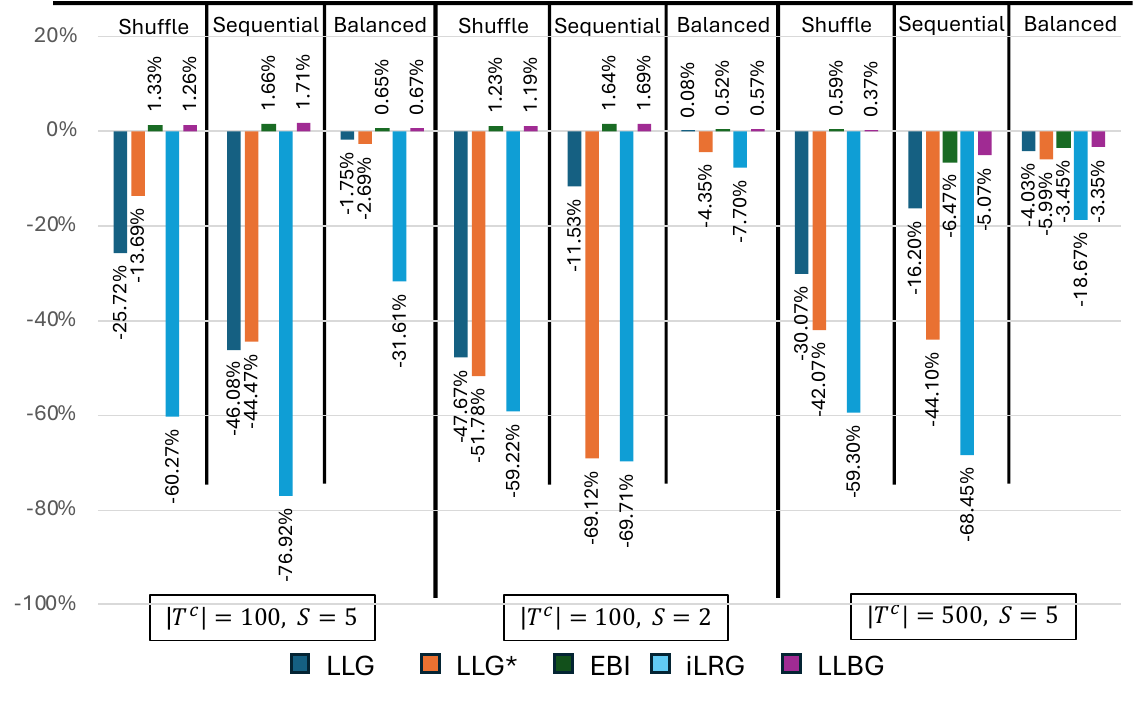}
    \caption{Exemplary multi-step experimental results applied on the Wetlab dataset for an untrained DeepConvLSTM. We report the average LnAcc difference between single step and multi-step gradient updates using different amounts of client data ($|T^c|$) and local update steps ($S$) for multi-step and fixing single-step experiments at $|T^c|=100$ and $S=1$. Positive values indicate increased leakage in multi-step updates, while negative values indicate reduced leakage.}
    \label{fig:multistep}
    \Description{Exemplary multi-step experimental results applied on the Wetlab dataset for an untrained DeepConvLSTM. We report the average LnAcc difference between single step and multi-step gradient updates using different amounts of client data ($|T^c|$) and local update steps ($S$) for multi-step and fixing single-step experiments at $|T^c|=100$ and $S=1$. Positive values indicate increased leakage in multi-step updates, while negative values indicate reduced leakage.}
\end{figure}

\paragraph{Multi-step gradient updates}

Since communication costs between clients and server are a critical resource in federated learning, McMahan et al. \cite{mcmahanCommunicationEfficientLearningDeep2017} proposed that clients perform multiple local gradient update steps before communicating with the server. This approach, known as FedAVG, significantly reduces communication overhead, as clients no longer need to transmit updates after every local step. In the context of our investigation into label leakage attacks, we again assume that the client and server agree on a fixed number of data points, $|T^c|$, for local training before sending an update. We explore whether averaging gradients over $S$ local updates, i.e., splitting $|T^c|$ into $S$ equal-sized mini-batches, can reduce label leakage risks from a single user.

Figure~\ref{fig:multistep} shows sample results comparing the difference in performance between single-step gradient update experiments ($|T^c|=100; S=1$) with multi-step variants. Results show that bias-based attacks, with the exception of iLRG, maintain stable performance even when reconstructing multi-step updates from untrained models. When applied to trained models however, LLBG and EBI exhibit more significant reductions in leakage, particularly under balanced sampling. Notably, an increased $|T^c| = 500$ results in the largest performance drop in ClassAcc and LnAcc for sequential sampling, which we hypothesize is due to batches becoming less class-exclusive as more data points are included, thereby reducing the effectiveness of reconstruction. Interestingly, iLRG, while heavily affected by multi-step updates, proves to be the more stable when applied to a trained models, and even increases in performance in certain cases when applied to shuffled and balanced batches. Finally, our results indicate that single-step and multi-step updates pose similar privacy risks. Though larger amounts of local client data showed to decrease leakage, it does not hide users label information effectively with LnAcc and LeAcc still remaining high, suggesting certain classes still being able to be reconstructed. 

\begin{figure*}
    \centering
    \includegraphics[width=0.72\linewidth]{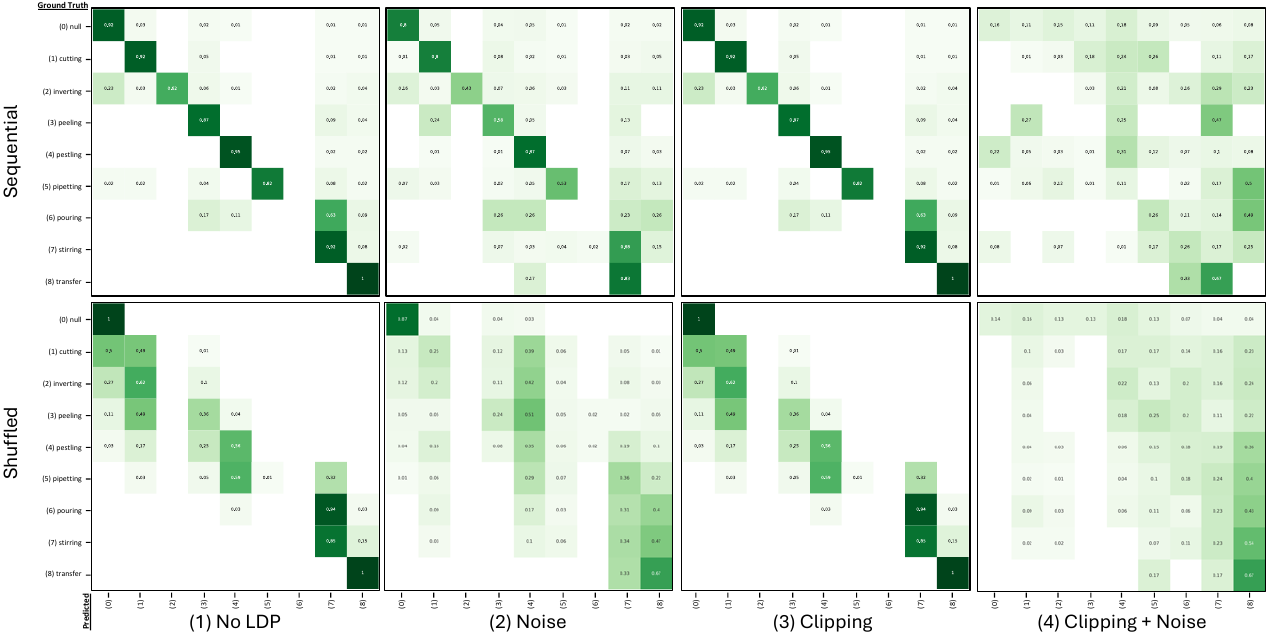}
    \caption{Results of the LLBG label leakage attack being applied to the Wetlab dataset using a trained DeepConvLSTM. Confusion matrices show the reconstructed label accuracy for the shuffled and sequential sampling case when using (1) no local differential privacy, (2) gaussian noise addition (3) gradient clipping and (4) the combination of both as described in the paper.}
    \label{fig:ldp}
    \Description{Results of the LLBG label leakage attack being applied to the Wetlab dataset using a trained DeepConvLSTM. Confusion matrices show the reconstructed label accuracy for the shuffled and sequential sampling case when using (1) no local differential privacy, (2) gaussian noise addition (3) gradient clipping and (4) the combination of both as described in the paper.}
\end{figure*}

\paragraph{Single-step with LDP}

Having shown that multi-step updates alone do not sufficiently mitigate label leakage, we further investigate whether LDP techniques, specifically gradient clipping and gradient noise, can better protect clients’ label information from the server. We repeated our initial experiments, this time applying one of the following LDP configurations: (1) gradient noise (Gaussian noise with mean 0 and standard deviation 0.1), (2) gradient clipping (scaling gradient vectors to an L2-norm of 0.1), or (3) a combination of both. Importantly, LDP measures were applied only to the gradients of the last model layer, as these are targeted by the label leakage attacks. Figure~\ref{fig:ldp} presents representative results for the LLBG attack applied to a trained DeepConvLSTM model. Overall, when comparing across all attack methods, we observe that gradient clipping alone is largely ineffective, particularly on gradients from sequentially sampled batches. With the exception of iLRG, other methods exhibit minimal change in effectiveness with clipping applied. Among all techniques, EBI remains the most stable, maintaining consistent performance across all LDP variants.

Sequential sampling continues to be the most vulnerable configuration, with EBI and LLBG achieving more than 70\% ClassAcc for untrained models and more than 60\% for trained models, even in the presence of noise. Nevertheless, noise addition significantly reduces LnAcc across all sampling methods, suggesting less precise label prediction. However, LeAcc remains high, indicating that attackers can still reliably infer whether a label is present in the batch. Notably, only the combination of clipping and noise proves to be an effective defense under sequential sampling. While this combined approach offers meaningful defense against bias-based attacks on sequential batches, LeAcc scores still exceed 50\% across all leakage attacks and even up to 80\% in case of the iLRG attack on the Wetlab dataset. This indicates that the presence of classes can still be recovered with relatively high accuracy, raising persistent privacy concerns despite the application of LDP and underscoring the need for more robust privacy-preserving strategies, especially for protecting less frequent activity classes.

\section{Limitations}
\label{sec:limitations}

Our paper presents the first comprehensive benchmarking of popular label leakage attacks on HAR benchmark datasets. While the effectiveness of these methods varies across architectures and datasets, our findings reveal a consistent trend: the unique characteristics of HAR data introduce significant privacy vulnerabilities that warrant serious attention. In particular, the inherent class imbalance in HAR datasets limits the reliability of conventional metrics such as LnAcc and LeAcc, commonly used in computer vision. These metrics tend to favor majority classes and can thus give a misleading impression of attack effectiveness. To address this, we introduced ClassAcc, a class-averaged metric that is calculated on a dataset-level, which contributes to a more class-balanced view of exhibited label leakage. Our evaluation of LDP defenses showed that while gradient noise and clipping are limited in isolation, their combination offers a more effective defense, yet certain attacks still achieved high recognition rates of class presence. Although more advanced privacy-preserving mechanisms exist, we chose these techniques due to their widespread adoption and computational simplicity and leave it to future work to explore more robust LDP strategies that can ensure privacy protection also in case of class presence. 

Additional experiments (provided in the code repository) further indicate that applying clipping and noise to the gradients of the last layer does not impair the prediction accuracy of trained models. As our experiments focus on single-client gradient updates, future research could examine how multi-client aggregation impacts label leakage, particularly in real-world FL deployments. Additionally, since bias-based attacks consistently outperformed weight-based ones, a practical defense recommendation may be to train HAR models without bias terms, as suggested in \cite{scheligaCombiningStochasticDefenses2022}. Further investigation could also be warranted into why iLRG was the most volatile among all label leakage attacks studied, with its performance highly sensitive to LDP measures and multi-step averaging. Moreover, although commonly employed in HAR tasks, we deliberately did not apply a weighted loss during training. Weighted losses increase the influence of rare classes on the gradient, potentially making these classes more susceptible to leakage, an especially critical concern in sensitive domains such as healthcare and disease detection. Lastly, given the scope of our experimental framework, we report only the most representative trends in the paper. To support reproducibility and future research, we have open-sourced all code and experiments in our code repository. 

\section{Discussion \& Conclusions}
\label{sec:conclusion}
This paper presented a first-of-its-kind analysis of label reconstruction in federated HAR. A comprehensive evaluation of five gradient inversion techniques demonstrated that the unbalanced nature of HAR datasets poses significant privacy risks that can be exploited by leakage attacks. LDP measures, namely gradient clipping and noise addition, proofed to be partially effective when combined, but insufficient to ensure full privacy in cases involving sequential sampling. Regardless of whether leakage methods were applied to multi-step averaged gradients or LDP-distorted gradients, label-exclusive batches still allowed certain attacks to infer the presence of classes with high accuracy. Our analysis suggests that the number of activity classes, the degree of class imbalance, and the sampling strategy are the main factors influencing the extent of label leakage.

Applying this to a real-world setting, we see a significant risk to stream-based HAR applications, such as those found in wearable fitness trackers, smart homes, or healthcare monitoring, where data is processed and uploaded continuously. In such settings, users often send gradient updates immediately after short recording sessions. If these updates are based on label-exclusive batches (i.e., containing only one activity), they become especially vulnerable to label leakage. To mitigate this, system designers should implement mechanisms that delay model updates until a sufficient and diverse amount of data has been collected, ideally covering all or most activity classes. This allows for the use of shuffled sampling strategies, which are significantly more robust against leakage. Moreover, applying stronger privacy protections, such as combining gradient clipping with noise addition, should be considered mandatory when sequential or imbalanced sampling from a small number of classes cannot be avoided.

\begin{acks}
We gratefully acknowledge the DFG Project WASEDO (grant number 506589320) and the University of Siegen's OMNI cluster.
\end{acks}

\bibliographystyle{ACM-Reference-Format}
\balance
\bibliography{main}

\end{document}